\def\year{2020}\relax
\documentclass[letterpaper]{article} % DO NOT CHANGE THIS
\usepackage{ai20}  % DO NOT CHANGE THIS
\usepackage{times}  % DO NOT CHANGE THIS
\usepackage{helvet} % DO NOT CHANGE THIS
\usepackage{courier}  % DO NOT CHANGE THIS
\usepackage[hyphens]{url}  % DO NOT CHANGE THIS
\usepackage{subfigure}
\usepackage{amsmath}
\usepackage{multirow}
\usepackage{booktabs}
\usepackage{graphicx} % DO NOT CHANGE THIS
\usepackage{graphicx}  % DO NOT CHANGE THIS
\title{ AVR: Attention based Salient Visual Relationship Detection }
\author{\Large \textbf{Jianming Lv, Qinzhe Xiao, Jiajie Zhong}\\ % All authors must be in the same font size and format. Use \Large and \textbf to achieve this result when breaking a line
South China University of Technology, Guangzhou, China\\ %If you have multiple authors and multiple affiliations
jmlv@scut.edu.cn, qinzhexiao@gmail.com, cszhongjiajie@mail.scut.edu.cn % email address must be in roman text type, not monospace or sans serif
}

\begin{document}

\maketitle

\begin{abstract}
   Visual relationship detection aims to locate objects in  images and recognize the relationships between objects. Traditional methods treat all observed relationships in an image equally, which causes a relatively poor performance in the detection tasks on complex images with abundant visual objects and various relationships. To address this problem, we propose an attention based model, namely AVR, to achieve salient visual relationships  based on both local and global context of the relationships. Specifically, AVR recognizes relationships and measures the attention on the relationships in the local context of an input image by fusing the visual features, semantic and spatial information of the relationships. AVR then applies the attention  to assign important relationships with larger salient weights for effective information filtering. Furthermore, AVR is integrated with the priori knowledge in the global context of image datasets to improve the precision of relationship prediction, where the context is modeled as a heterogeneous graph to measure the priori probability of relationships based on the random walk algorithm. Comprehensive experiments are conducted to demonstrate the effectiveness of AVR in several real-world image datasets, and the results show that AVR outperforms state-of-the-art visual relationship detection methods significantly by up to $87.5\%$ in terms of recall.
\end{abstract}

\section{Introduction}
As a critical task of scene understanding, visual relationship detection aims to identify the objects in an image, and  recognize the relationship between each pair of objects. The visual relationship can be represented by a triplet $<$subject, predicate, object$>$ where the predicate is the semantic interaction between the subject and object. The interaction can be spatial relationships (e.g., on, under and behind) or verbs (e.g., eat, walk on and play). The detected visual relationships are useful structured information that can be used by many other high-level applications such as image retrieval \cite{ImageRetrieval}, image captioning \cite{ImageCaptioning}, and visual question answering \cite{R-VQA}.

\begin{figure}[t]
   \centering
   \includegraphics[width=\linewidth]{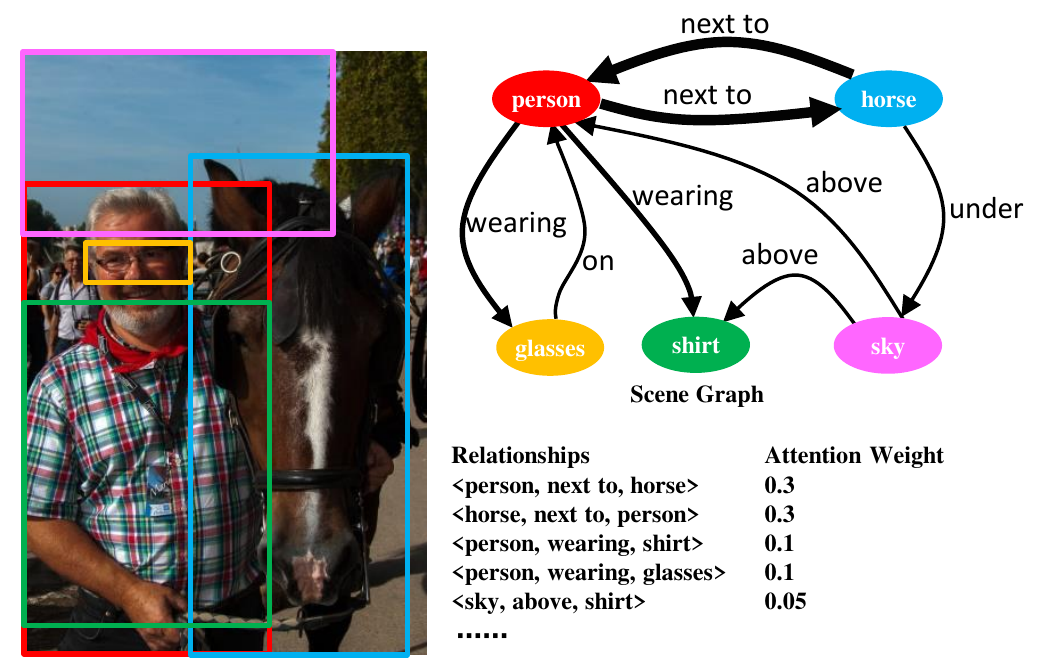}
   \caption{Attention on Salient Visual Relationships. From human perspective, we usually pay more attention on relatively salient and meaningful relationships among abundant relationships in an image. For example, the relationships $<$person, next to, horse$>$ and $<$horse, next to, person$>$ are more important and salient than the other relationships such as $<$sky, above, shirt$>$. }
   \label{fig:1}
\end{figure}

It is a great challenge to detect meaningful relationships from natural scene images, which usually contain numerous visual objects and complex relationships between objects. Most of existing algorithms \cite{LangPrior}, \cite{LKDistill}, \cite{CAI}, \cite{MessagePassing} divide this task into two basic stages:  detecting objects, and recognizing the predicate of each pair of detected objects. The detected relationship triplet and the bounding boxes of objects are the final outputs of the algorithms. On the other hand, since the space of all possible relationships is extremly huge and the training dataset hardly cover all possible combinations of objects and predicates. The ability of few-shot and even zero-shot learning is quite necessary to enhance the relationship detection model. Recently, some priori information based methods are proposed to address this problem. Lu et.al. \cite{LangPrior} propose a language model to predict the priori probability of predicates between objects, and then apply the priori knowledge  to finetune the predicate probability predicted from the visual model. \cite{LKDistill} integrates priori knowledge from the statistics of language text (Wikipedia) so as to optimize the prediction of unseen relationships.  On the other hand, the literatures \cite{CAI}, \cite{DRnet}, \cite{MessagePassing} utilize the context information to help to predict the relationships better. The context can be the objects \cite{CAI}, \cite{DRnet} or other relationships \cite{MessagePassing} in an image.

Most of the above algorithms focus on correctly detecting all of the relationships in an image. However, from the human perspective, the relationships in an image usually have different importance and we usually pay attention to a small portion of salient and meaningful relationships as demonstrated in Fig.1. Simply treating all observed relationships  equally and independently may cause a relatively poor performance in the detection tasks on complex images with abundant objects.
% !!!!!!
% todo: replace Fig.1. with ref
% !!!!!!

 Motivated  by the observation above, we propose an attention based algorithm AVR to take into account both of local and global context of a relationship for precise detection of salient visual relationships. In particular, AVR measures  the attention on each visual relationship based on its visual features, semantic and spatial information in the local context of the input image.  The salient relationships, which are assigned with larger attention, indicate important semantic  of the images from human perspective. Moreover, AVR utilizes the priori knowledge in the global context of image datasets to further improve the performance of identifying correct relationship. Specifically, the priori knowledge is modeled as a heterogeneous graph, which takes all possible predicates and object pairs as vertices and utilizes the random walk algorithm \cite{RandomWalk} to propagate the priori probability of relationships.

 As validated in the comprehensive experiments on the real-world image datasets, AVR outperforms state-of-the-art visual relationship detection methods significantly by concentrating on salient relationships accurately.
 
In short, our major contributions are the following:
\begin{itemize} 
    \item We propose a novel attention based visual relationship detection algorithm AVR to detect the relationships and pay more attention on salient relationships by fusing the visual features, semantic and spatial information in the local context of the input image.

    \item We model the priori knowledge as a heterogeneous graph in the global context of image datasets, and apply random walk algorithm to capture the priori probability of relationships, which is integrated into the detection model to enhance the precision of predicate prediction.
    
    \item Comprehensive experiments on the commonly used dataset VRD \cite{LangPrior} and VG \cite{VGDataset} are conducted to test the performance of AVR, and the results show that our algorithm can significantly outperform  state-of-the-art methods with the improvements by up to $87.5\%$ in terms of recall.
    
\end{itemize}   
    
The rest of this paper is organized as follows: In
Section \ref{sec:related_work}, we present a brief literature review which is related to our work. In Section \ref{sec:AVR}, we introduce our proposed model AVR for visual relation detection. The experiments are presented in Section \ref{sec:experiments}. Finally,  the conclusion is given in Section \ref{sec:conclusion}.

\section{Related Work}\label{sec:related_work}

In the visual relationship detection, recognizing the interaction between the objects is a primary task, while localizing objects in an image are generally done by the existing popular detection algorithms (e.g., Faster R-CNN \cite{Faster-RCNN} and YOLO9000 \cite{YOLO9000}). As mentioned before, the relationship distribution is long-tailed and the training examples of many of relationships are limited. Therefore, some studies proposed to utilize the language priori knowledge of relationships to improve the few-shot learning ability of relationship prediction. Lu et al. \cite{LangPrior} proposed a language model to predict the predicate priori probabilities by utilizing the semantic word vectors of objects. Then the priori probabilities are used to finetune the relationship prediction of visual model. Different from \cite{LangPrior}, Yu et al. \cite{LKDistill} obtained the priori distribution by the statistics from collected Wiki textual data, and integrated the obtained priori into the prediction model. The priori from textual data is more precise, but it needs extra collection and processing for textual data. 
% In addition, Hwang et al. \cite{Tensorize} yielded the priors through building a tensor representing the appearing times of relationships, and factoring the tensor to get a dense tensor.

Context information is another direction studied by researches. The literatures \cite{CAI}, \cite{DRnet}, \cite{ViP-CNN}, \cite{Zoomnet} attempted to predict objects and predicates with the context of relationships. Danfei Xu et al. \cite{MessagePassing} proposed to use all relationships in an image as context to help the predicate prediction, and the context is propagated by iterative message passing in a graph network. Li et al. \cite{FactorizableNet} proposed the Factorizable Net which also propagates context in a graph, but the graph is simplified by merging and clustering some repeated edges for speedup. Guojun Yin et al. \cite{Zoomnet} proposed a spatiality context appearance module to put the relative position information between objects and predicate into consideration.

There are many other types of researches on visual relationship representation. For example, VtransE \cite{Vtranse} applied the representation of relationship in knowledge graph \cite{KnowledgeGraphEmb} to the visual relationship representation. Yang et al. \cite{Shuffle-then-assemble} improved the generalization of model through learning object-agnostic visual features. Zhu et al. \cite{structuredLearning} proposed a deep structured model for learning relationship in both feature-level and label-level prediction which can capture the dependencies between objects and predicates. Liang et al. \cite{StructuralRanking} proposed a structural ranking objective function to assign higher scores to annotated relationships than unannotated relationships.

% (desc of VtransE \cite{Vtranse}) Specifically, in the relationship space, the predicate vector is equal to subject vector subtracted by object vector.

\section{The AVR Model}\label{sec:AVR}

\begin{figure}[t]
   \centering
   \includegraphics[width=\linewidth]{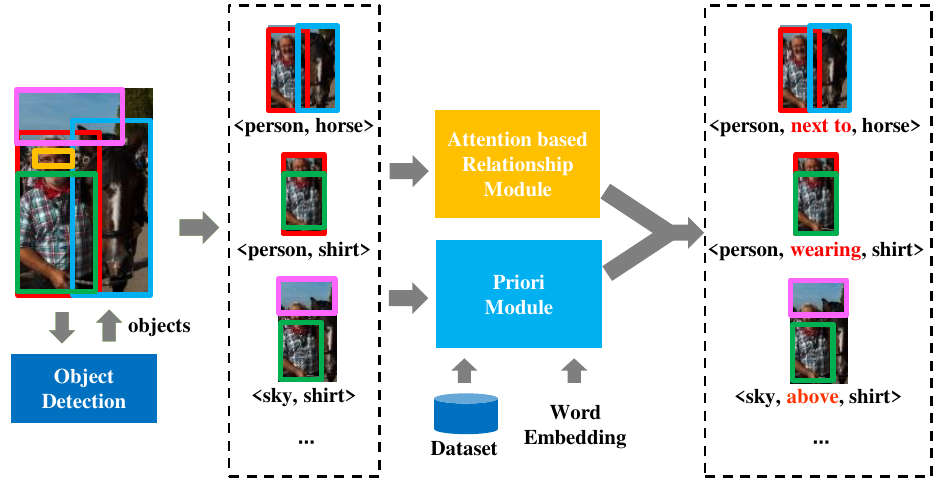}
   \caption{The framework of the AVR model. }
   \label{fig:framework}
\end{figure}

Fig. \ref{fig:framework} shows the framework of the proposed AVR model. Given an input image, the object detection module is firstly applied to extract the candidate objects, each one of which is represented as a class label and a bounding box in the image.  Each pair of objects is then fed into the Attention based Relationship Detection Module, which makes decision based on the local context of the input image, to predict the probability of  the relationship represented in the objects. Meanwhile, based on the modeling of the global context related to the image datasets, the priori knowledge about relationship distribution is also integrated into the Priori Module to refine the accuracy of relationship detection. The details of each module are given as follows.

% Given an image, the object detector detects the objects. Then each pair of objects is passed through the attention based relationship prediction module and priori module to obtain the salient relationships in the image.

%Given an image, the detection model will detect the class and bounding boxes of objects in the image. Then, for each pair of objects, the proposed attention model will recognize their predicate and output a weight w representing the importance of this relationship. The features used for predicting predicate are semantic feature, spatial feature and visual feature, which can improve the robustness of model. In the attention module, we use the whole image conv feature to extract image feature as context, which is connected with the visual feature of a relationship to determine the salient weight of the relationship. In the prior extraction module, we use statistic of the training data and the random walk to obtain better prior distribution of the relationship. Finally, the final predicate prediction is decided by predicate probability, salient weight and prior distribution. The more details of each module are given as follows.

\subsection{Attention based Relationship Prediction}

Like most of the visual relationship detection algorithms \cite{CAI}, \cite{LKDistill}, \cite{Vtranse}, we adopt the object detection algorithm, Faster R-CNN \cite{Faster-RCNN}, to retrieve all objects in the input image. Each detected object $O_j$ is represented as a tuple : $(B_j, V_j,C_j, Pr(C_j|O_j))$. $B_j$ is the bounding box of the object, and is presented as a tuple $(x_j, y_j,w_j,h_j)$, where $(x_j, y_j)$ indicates the left-top position of the object in the image, $w_j$ and $h_j$ are the width and height of the object. While apply $B_j$ to crop the input image, we can achieve the object image, which is denoted as $V_j$. $C_j$ is the most possible class label of the object. $Pr(C_j|O_j)$ means the confidence for the label $C_j$.

After obtaining a list of detected objects$\{O_j\}$, the combination of any pair of objects  can constitute a latent relationship in an image.  The key of scene understanding is to achieve the most  salient visual relationships from these candidates, which represent the most important semantic in the image.

\subsubsection{Bayesian Network based Relationship Inference}

\begin{figure}[t]
   \centering
   \includegraphics[width=0.5\columnwidth]{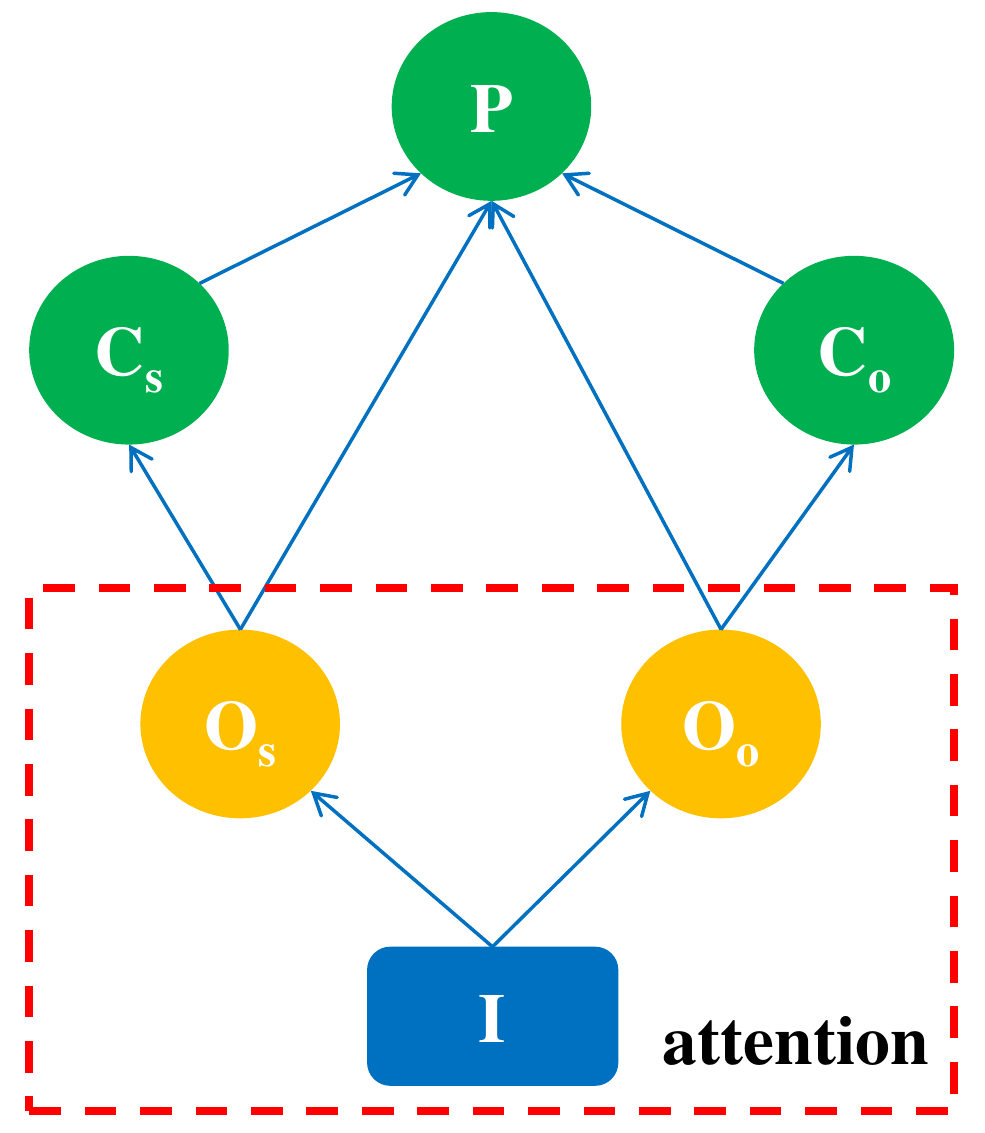}
   \caption{  Bayesian Network based Relationship Inference.} 
   \label{fig:bayesian}   
\end{figure}

Inspired by the human information processing procedure, we propose an attention based relationship prediction model, which considers visual relationship detection as an  inference procedure in a Bayesian network as Fig.~\ref{fig:bayesian}. Specifically, given an input image $I$, two objects are detected and selected as the subject  $O_s$ and object  $O_o$ of the visual relationship. $C_s$ and $C_o$ are the class labels of the selected subject and object respectively. The predicate of the visual relationship $P$ is inferred based on the subject $O_s$, the object $O_o$ and their class labels ( $C_s$, $C_o$). According to the Bayesian chain rule, the inference probability of a relationship conditioned on the observed image $I$ can be formulated as follows:
    \begin{eqnarray}
        Pr (P,C_s,C_o,O_s,O_o| I) = Pr (O_s, O_o|I) \cdot  \nonumber \\ 
        Pr(C_s|O_s)\cdot Pr(C_o|O_o) \cdot Pr(P|C_s, C_o, O_s, O_o)     
    \label{eq:bayesian}   
    \end{eqnarray}  
Here $Pr (O_s, O_o|I)$ indicates the probability of attention on the object pair $(O_s, O_o)$ , which is selected from the image $I$. $C_s$ is the most possible class label of $O_s$, the probability of which is $Pr(C_s|O_s)$. Similarly, $C_o$ is the most possible class label of $O_o$, the probability of which is $Pr(C_o|O_o)$. Both of  $Pr(C_s|O_s)$ and $Pr(C_o|O_o)$ are given by the object detection module. $Pr(P|C_s, C_o, O_s, O_o)$ means the probability of the predicate, which is predicted based on the detected subject and object. The detail of calculating the predicate probability $Pr(P|C_s, C_o, O_s, O_o)$ and the attention probability $Pr (O_s, O_o|I)$  will be introduced in the following sections.

 \subsubsection{Predicate Prediction Module} \label{sec: predicate-prediction}

 \begin{figure*}[t]
   \centering
   \subfigure[]{
       \includegraphics[width=1\columnwidth]{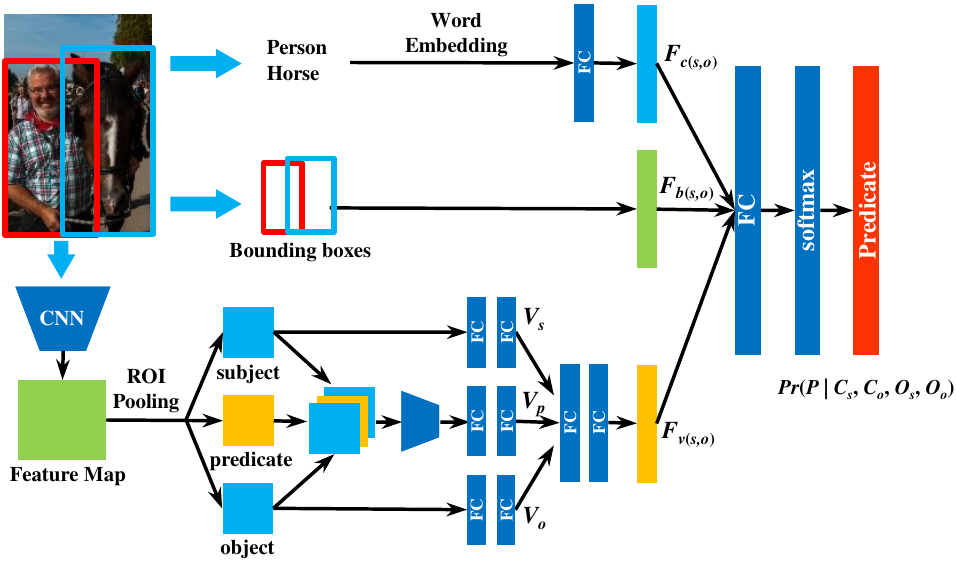}
       \label{fig:attention-detection-a}
   }
   \hspace{.3in}
   \subfigure[]{
      \centering
       \includegraphics[width=0.85\columnwidth]{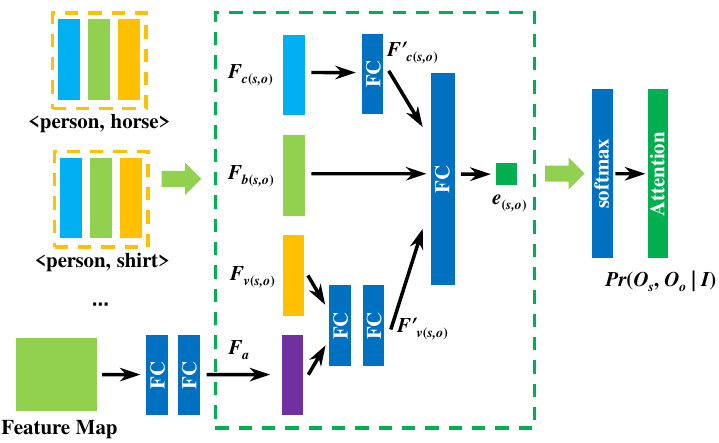}
       \label{fig:attention-detection-b}
   }
   \caption{\textbf{The Attention based Relationship Prediction}: (a) \textbf{Predicate Prediction Module} for recognizing predicates between subjects and objects; (b) \textbf{Attention Module} for measuring the attention weights of relationships in an image. Both of modules utilize and fuse visual, spatial and semantic features of objects. }
   \label{fig:attention-detection}
\end{figure*}

In order to achieve the predicate prediction result $Pr(P|C_s, C_o, O_s, O_o)$, we construct the multi-modal fusion model which integrates the visual features, spatial information and semantic feature of detected objects as shown in Fig.~\ref{fig:attention-detection-a}.  Given any pair of detected objects $O_s:(B_s, V_s, C_s, Pr(C_s|O_s))$ and $O_o:(B_o, V_o, C_o, Pr(C_o|O_o)$, the bounding box containing these two objects is defined as the predicate bounding box $B_p$, which is the smallest rectangle containing both of $B_s$ and $B_o$.  The image within the bounding box  $B_p$ is called the predicate image $O_p$. Similar to Faster R-CNN \cite{Faster-RCNN}, by applying the bounding boxes on the feature maps output by the CNN model, we can achieve the feature maps of $O_s$, $O_o$ and $O_p$. Subsequently, the ROI pooling operation is applied to transform these feature maps into the ones the same size. Then the feature maps of  $O_s$ / $O_o$ are passed to a two fully connected layers to obtain the subject/object feature $V_s$ / $V_o$. Meanwhile, the feature maps of  $O_s$, $O_o$ and $O_p$ are stacked and processed by the following three convolutional layers and two fully connected layers to achieve the predicate feature vector $V_p$. Finally, $V_p$ is concatenated with $V_s$ and $V_o$ together and fed into the two fully connected layers to get the final visual feature $F_{v(s,o)}$,  which contains the visual information of the subject, object and their local context. 

Besides the visual feature, the spatial feature is also integrated into the prediction model as shown in Fig.~\ref{fig:attention-detection-a}. Particularly, given the bounding box of the subject $B_s = (x_s, y_s, w_s, h_s)$, its normalized bounding box is $[x_s/W, y_s/H, (x_s+w_s)/W, (y_s+h_s)/H, A_s/A_I]$, where $W$ and $H$ are the width and height of the image, $A_s$ is the area of the subject, and $A_I$ is the area of the input image. Similarly, the normalized bounding box of the object $O_o$ is  $[x_o/W, y_o/H, (x_o+w_o)/W, (y_o+h_o)/H, A_o/A_I]$, where  $A_o$ is the area of object.  The relative position feature vector of these two boxes is $ [(x_s-x_o)/W, (y_s-y_o)/H, log(w_s/w_o), log(h_s/h_o)]$. This relative position feature vector is concatenated with the normalized bounding boxes of the subject and object to form the complete spatial feature $F_{b(s,o)}$. 

Furthermore, the class labels $C_s$ and $C_o$ are also utilized to support precise prediction. The word embedding technique \cite{word-vector} is utilized to represent these labels as embedding vectors, which are concatenated and input into the full connected layers to obtain the semantic feature $F_{c(s,o)}$. At last, the visual feature $F_{v(s,o)}$, the spatial feature $F_{b(s,o)}$ and the semantic feature $F_{c(s,o)}$ are fused in the softmax layer to achieve the final prediction probability as follows:
\begin{eqnarray}
   Pr(P|C_s, C_o, O_s, O_o)= \nonumber \\
   softmax( W_vF_{v(s,o)} + W_cF_{c(s,o)} + W_bF_{b(s,o)} + b)
\end{eqnarray}
where $W_v$, $W_c$, $W_b$ and $b$ are the parameters in the model.

While training the model, the following cross entropy loss function is adopted to measure the precision of prediction:
\begin{equation}
   \begin{split}
   Loss_P = \sum_{(s,o)} \sum_i -y_i log(Pr(P_i|C_s, C_o, O_s, O_o))  
   \end{split}
   \label{eq-loss-p}
   \end{equation}
Here $y_i$ is the indicator of ground truth predicate label. $y_i$ is equal to 1 if the $i$th predicate is the ground truth label of the $O_s$ and $O_o$, otherwise $y_i$ is equal to 0. 

\subsubsection{Attention Module}
As mentioned in Eq.~(\ref{eq:bayesian}), the attention module aims to measure the probability of focusing on a pair of objects in an image: $Pr (O_s, O_o|I)$, which indicates the attention on the selected objects.  Inspired by the human attention mechanism,  we measure the attention based on the visual, spatial, and semantic information of the objects as shown in Fig.~\ref{fig:attention-detection-b}. Specifically, while considering the visual clues, we combine the visual feature of the objects $F_{v(s,o)}$ and the convolutional feature $F_a$ of the whole image in the following manner:
\begin{equation}   
   F_{v(s,o)}' = R(W_3 R(W_1 F_{v(s,o)} + W_2 F_a +b_1)+b_2)
\end{equation}
,where $W_i$ and  $b_i$ ($i \in Z^{+}$) are all tunable parameters in the nerual network. $R(.)$ is the nonlinear activation function Relu. Furthermore, as shown in Fig.~\ref{fig:attention-detection-b}, the spatial feature $F_{b(s,o)}$ and the semantic feature $F_{c(s,o)}$ are integrated with the visual feature $ F_{v(s,o)} $ to achieve the attention score as:
\begin{equation}
   e_{(s,o)} =  R(W_4 F_{v(s,o)}' + W_5 F_{b(s,o)} + W_6 F_{c(s,o)}' +b_3  ) 
\end{equation}
,where $F_{c(s,o)}'$ is the a nonlinear transformation of the semantic feature $F_{c(s,o)}$:
\begin{equation}
   F_{c(s,o)}' =  R(W_7 F_{c(s,o)} +b_4) 
\end{equation}

Finally, the softmax funcion is applied to achieve the normalized attention: 
\begin{equation}
   Pr (O_s, O_o|I) = \frac{exp(e_{(s,o)})}{\sum_{i,j}exp(e_{(i,j)})}   
\end{equation}

When training the attention module, we regard it as a binary classification problem to predict whether a relationship is important or not from the human perspective. The  cross entropy loss for each input image is defined for training as follows.
\begin{equation}
Loss_A = \sum_{(s,o)} (-L_{(s,o)} log(\sigma(e_{s,o})) - (1-L_{(s,o)}) log(1-\sigma(e_{s,o}))
\end{equation} 

Here $L_{(s,o)}$ indicates the attention label of the objects pair $O_s$ and $O_o$. Since there are not annotated importance of relationships on datasets, we simply regard the annotated relationships are important than those unannotated. Thus, if $O_s$ and $O_o$ are the subject and object of a annotated relationship on dataset, $L_{(s,o)}$ is 1. Otherwise, $L_{(s,o)}$ is 0.  $\sigma$ is the sigmoid function transforming the attention score into the scope $[0,1]$. This attention loss is combined with the predicate loss of Eq.(\ref{eq-loss-p}) in the following form to optimize the whole model by minimizing the loss.
\begin{equation}
Loss = Loss_A + Loss_P
\end{equation}

\subsection{Priori Knowledge Graph based Enhancement}\label{sec:priori}
\begin{figure}[t]
   \centering
   \includegraphics[width=0.8\columnwidth]{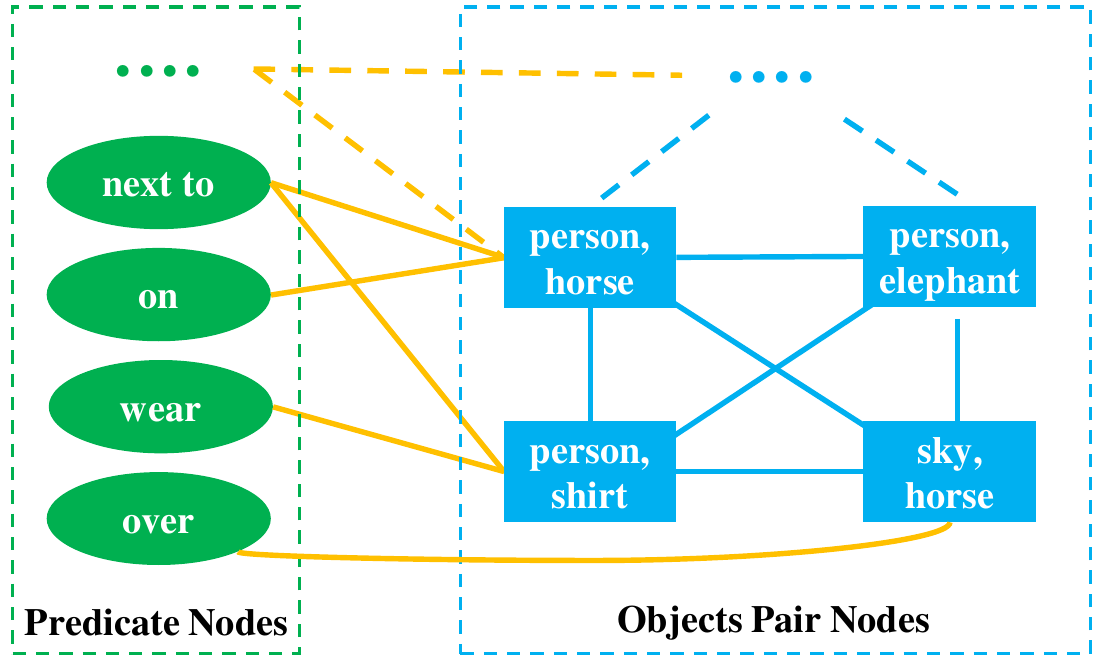}
   \caption{The heterogeneous graph of relationships.} 
   \label{fig:heterogeneous}   
\end{figure}

We also integrate the priori knowledge in the global context of the whole dataset to further improve the precision of relationship detection. Specifically, we model the priori knowledge as a heterogeneous graph (shown in Fig. \ref{fig:heterogeneous}), where all possible predicates and object pairs are denoted as nodes. The edge connecting any predicate $P_i$ and any object pair $(C_s, C_o)$ indicates the relationship $<$$C_s, P_i, C_o$$>$, which is labeled in the datasets.  The weight of the edge indicates the frequency that the relationship appears in the dataset. However, the graph is usually very sparse because of the long-tailed relationship distribution. Thus, we augment the graph by adding the edges between objects pair nodes to measure their similarity, where the weight of each edge is assigned with the similarity of the embedding vectors of objects. 

Based on this augmented heterogeneous graph, we infer the dependency of predicates and object pairs  by performing random walk from predicates to object pairs.  In this way, the priori probability $Pr(C_s,P_i, C_o)$ can be measured by the probability of that the random walker starting from $P_i$ reaches the object pair $(C_s, C_o)$.

Specifically, the adjacency matrix between predicates and object pairs are denoted as $D_0$ with size $K \times N^2 $, where $N$ is the number of object categories and $K$ is the number of predicate categories.  Furthermore, the adjacency matrix between object pairs is  an $N^2\times N^2$  matrix $M$. Both $D_0$ and $M$ is normalized by rows. The transition probability matrix of the $t$-step random walk from predicates to object pairs is: 
\begin{equation}
D_{t+1} = D_{t} M
\end{equation}
,where $t$ indicates the $t$-th iteration of the random walk.  Following the research on random walk based dependency inference \cite{RandomWalk-reid}, we add a balance parameter $\lambda$ to  prevent the updated $D_{t+1}$ deviates too far from initial $D_0$.
\begin{equation}
D_{t+1} = \lambda D_tM + (1-\lambda)D_0   
\end{equation}
When $t$ tends to infinity, the final  priori probability is
\begin{equation}
Pr(C_s,P_i, C_o) = D_{\infty} = (1-\lambda)(I - \lambda M)^{-1} D_0   
\label{eq:priori}
\end{equation}

By combining Eq.~(\ref{eq:bayesian}) and Eq.~(\ref{eq:priori}), we can achieve the final decision of a relationship $<$$C_s,P, C_o$$>$ in the image $I$:
\begin{equation}
f(C_s,P, C_o,I) = Pr (P,C_s,C_o,O_s,O_o| I) \cdot Pr(C_s,P, C_o)
\label{eq:fianl-score}
\end{equation} 
,which combines both of posterior  probability and priori knowledge in the datasets for precise prediction.

\section{Experiments} \label{sec:experiments}
 
\textbf{Datasets:} In the experiments, the VRD \cite{LangPrior} and VG \cite{VGDataset} datasets are used to verify the effectiveness of the proposed  AVR model. Specifically, VG  \cite{VGDataset} is a large image dataset  with over 2 million annotated relationships. It is pre-processed in different ways in  previous literatures \cite{Vtranse}, \cite{MSDN}, \cite{Zoomnet}. For fair comparison,  we use two commonly used versions of VG, i.e., VG-VtransE \cite{Vtranse} and VG-MSDN \cite{MSDN}. The details of the datasets are showed in Table \ref{tab:stat}. Here $\rm N_{obj}$ denotes the number of object categories and $\rm N_{pred}$ denotes the number of predicate categories. $\rm N_{img}$ denotes the number of images, while $\rm N_{rel}$ represents the average number of annotated relationships in an image.

\begin{table}[t]
\small
\centering
\renewcommand\tabcolsep{4.0pt}
\begin{tabular}{ccccccc}
\toprule
\multirow{2}{*}{Dataset}  & \multirow{2}{*}{$\rm N_{obj}$} & \multirow{2}{*}{$\rm N_{pred}$} & \multicolumn{2}{c}{Training Set} & \multicolumn{2}{c}{Test Set} \\
                          &     &     &  $\rm N_{img}$  & $\rm N_{rel}$ &  $\rm N_{img}$  & $\rm N_{rel}$ \\
\midrule
VRD      & 100 & 70  & 3780            & 8.03          & 954             & 8.00    \\
VG-VtransE  & 200 & 100 & 73794           & 9.23          & 25858           & 9.43    \\
VG-MSDN      & 150 & 50  & 46164           & 9.12          & 10000           & 9.17    \\
\bottomrule
\end{tabular}
\caption{The statistics of datasets.}
\label{tab:stat}
\end{table}

\textbf{Parameters Setting:} We use Faster R-CNN \cite{Faster-RCNN} to localize all objects in an image and VGG16 is selected as the backbone network to extract  visual features, because it is commonly used by recent literatures \cite{Zoomnet}, \cite{FactorizableNet}. The parameters of VGG16 are initialized by the parameters pre-trained in the Faster R-CNN on the training set, and kept unchanged during training the predicate prediction model. The optimization method we used is the SGD with momentum 0.9. In addition, the  embedding vectors of the caption words used to obtain the semantic feature $F_{c(s,o)}$  are the pre-trained vectors of Glove \cite{glove}, the dimension of which is 50. In the priori module of Section \ref{sec:priori}, the parameter  $\lambda$ of random walk is set to 0.5 for VRD and 0.3 for both VG-VtransE and VG-MSDN.

 \textbf{Visual Relationship Detection Tasks:} Generally, there are three kinds of popularly used tasks for visual relationship detection \cite{LangPrior}, which are listed as follows:

(1) \textbf{Predicate Detection:} This task aims to determine the predicate of  a given objects pair in an image, where the bounding boxes and class labels of the objects are provided.

(2) \textbf{Phrase Detection:} Given an image, this task aims to output a set of relationship triplets $<$subject, predicate, object$>$ in the image and localize each relationship with one bounding box. A detected relationship $<$subject, predicate, object$>$ is considered as a correct match if and only if its bounding box has at least 0.5 overlap with one of the  bounding boxes of  ground truth relationships.

(3) \textbf{Relationship Detection:} This task aims to output the relationship triplets $<$subject, predicate, object$>$ of a given image, and localize the bounding boxes of the subject and object of each relationship. If the bounding boxes of the subject and object have at last 0.5 overlap with the ground truth bounding boxes respectively, it is considered as a hit.

 \textbf{Metrics:} Following \cite{LangPrior}, \cite{Zoomnet}, the metric Rec@N used here is the recall rate of top $N$ predicted relationships, which are sorted by the probabilities of relationships output by the models. If multiple relationships with the same subject and object are detected by the model, only top $K$ relationships are selected in each group for testing. For example, if $K$ is 1,only the predicate  with maximum probability is selected for each pair of detected subject and object.

\subsection{Analysis of the Components in AVR}

\begin{table}[t]
    \small
    \centering
    \renewcommand\tabcolsep{1.0pt} % 调整表格列间的长度
    \begin{tabular}{ccccc}
    \toprule
    \multirow{2}{*}{Method} & \multicolumn{2}{c}{K=1} & \multicolumn{2}{c}{K=70} \\
       & Rec@50 & Rec@100 & Rec@50 & Rec@100 \\
    \midrule
    $\rm F_b$                        & 38.07 & 38.07 & 80.82 & 90.01 \\
    $\rm F_c$                        & 48.77 & 48.77 & 86.68 & 93.79 \\
    $\rm F_v$                        & 51.75 & 51.75 & 89.94 & 95.92 \\
    $\rm F_v$+$\rm F_b$                    & 52.36 & 52.36 & 90.35 & 95.99 \\
    Baseline($\rm F_v$+$\rm F_b$+$\rm F_c$)      & 54.54 & 54.54 & \textbf{91.47} & \textbf{96.65} \\
    $\rm Priori_{ds}$           & 51.59 & 51.59 & 82.51 & 87.65 \\
    $\rm Priori_{rw}$           & 51.87 & 51.87 & 88.71 & 94.30 \\
    Baseline+$\rm Priori_{ds}$  & 53.99 & 53.99 & 82.46 & 87.46 \\
    Baseline+$\rm Priori_{rw}$  & \textbf{55.61} & \textbf{55.61} & 90.73 & 95.72 \\
    \bottomrule
    \end{tabular}
    \caption{Evaluation of different variation models in the Predicate Detection task on VRD dataset.}
    \label{tab:feats_cmp}
    \end{table}

To analyze the effectiveness of different components of the AVR model, several variation models with different components are implemented and tested for comparison. The experimental results are illustrated on Table \ref{tab:feats_cmp} and Table \ref{tab:att_cmp}. $\rm F_b$ ($\rm F_c$ or $\rm F_v$) indicates the predicate prediction model of Fig.~\ref{fig:attention-detection-a} only using the spatial feature $F_b$ (semantic feature $F_c$ or visual feature $F_v$), while  $\rm F_v$+$\rm F_b$ represents the model using both visual feature and spatial feature. The Baseline ($\rm F_b$+$\rm F_c$+$\rm F_v$) is the predicate prediction model with all three features as shown in Fig.~\ref{fig:attention-detection-a} . For the priori module, the $\rm Priori_{ds}$ model uses the initial priori $D_0$ without random walk, and $\rm Priori_{rw}$ uses the final priori $D_{\infty}$ after random walk. Att means the attention module in the proposed AVR. Different combination of The Baseline, the priori module and the attention module are tested to verify the effectiveness of different components, e.g., Baseline+$\rm Priori_{rw}$ and Baseline+$\rm Priori_{rw}$+Att. Note that the attention module is not  used in the Predicate Detection task of Table \ref{tab:feats_cmp}, because the task aims to predict the predicate of given pair of subjects and objects in an image and it does not need to rank the relationships according to attention.

Table \ref{tab:feats_cmp} shows the results of the Predicate Detection task on the VRD dataset. It can be observed that the model $\rm F_v$ is better than $\rm F_b$ and $\rm F_c$, which indicates that relationships are more possible to be inferred from their visual appearance  than the spatial or semantic features. The Baseline model, which fuses the visual, spatial and semantic features, performs better than the models with one or two features. Besides, the $\rm Priori_{rw}$ obtains some improvement compared with the origin $\rm Priori_{ds}$, especially in the cases with larger $K$ where more possible predicates are allowed for output. Moreover, in the case where $K=70$, the recall rate of $\rm Baseline+\rm Priori_{rw}$ is surprisingly not better than the $\rm Baseline$ model. This may be because the VRD dataset is small-sized, and the priori knowledge extracted from the global context is not accurate enough. Actually, for the larger dataset VG, the $\rm Priori_{rw}$  can effectively improve the precision for all test cases as shown  in Table \ref{tab:cmp_vg}.

To further test the effectiveness of the attention mechanism of the AVR model, we verify the performance of the  attention module and the priori module in the Phrase and Relationship Detection tasks, which are much harder than the simple Predicate Detection task. The experimental results on the VRD dataset are listed in Table \ref{tab:att_cmp}. We can see that the Baseline model with the priori knowledge $\rm Priori_{rw}$ performs better than the Baseline for all tasks. Particularly, the attention module $\rm Att$ can further improve the performance significantly by up to 10\%$\sim$18\% for K=1 and 22\%$\sim$32\% for $K=70$. This confirms that the attention mechanism can effectively distinguish the importance of numerous relationships in an image and pick out the salient relationships for better scene understanding.

\begin{table}[t]
   \small
   \centering
   \renewcommand\tabcolsep{1.0pt} % 调整表格列间的长度
   \begin{tabular}{cccccc}
   \toprule
   \multirow{2}{*}{K} & \multirow{2}{*}{Method} & \multicolumn{2}{c}{Phrase Det.} & \multicolumn{2}{c}{Relaitonship Det.} \\
   & & Rec@50 & Rec@100 & Rec@50 & Rec@100 \\
   \midrule
   \multirow{4}{*}{1}   & Baseline                        & 23.25 & 28.25 & 17.53 & 21.08 \\
                        % & $\rm Prior_{rw}$           & 23.96 & 29.51 & 19.08 & 23.82 \\
                        & Baseline+$\rm Priori_{rw}$       & 24.46 & 30.21 & 19.15 & 23.15 \\
                        & Baseline+$\rm Priori_{rw}$+Att   & \textbf{29.33} & \textbf{33.27} & \textbf{22.83} & \textbf{25.41} \\
   \midrule
   \multirow{4}{*}{70}  & Baseline                        & 25.19 & 32.11 & 19.08 & 24.95 \\
                        % & $\rm Prior_{rw}$           & 25.27 & 33.70 & 20.17 & 26.91 \\
                        & Baseline+$\rm Priori_{rw}$       & 25.79 & 33.72 & 20.37 & 26.27 \\
                        & Baseline+$\rm Priori_{rw}$+Att   & \textbf{34.51} & \textbf{41.36} & \textbf{27.35} & \textbf{32.96} \\
   \bottomrule
   \end{tabular} 
   \caption{Evaluation of the $\rm Baseline$, $\rm Baseline+Priori_{rw}$ and $\rm Baseline+Priori_{rw}+Att$ in Phrase and Relationship Detection tasks on VRD dataset. } 
   \label{tab:att_cmp}
   \end{table}

\subsection{Comparisons with State-of-the-Art Methods}

We also compare AVR with several state-of-the-art methods and show the results on the VRD dataset in Table \ref{tab:cmp_vrd}. In the Predicate Detection task, AVR performs better than other methods except CAI+SCA-M \cite{Zoomnet} in the case K=1, but the AVR obtains much better performance in the case K=70 compared with the CAI+SCA-M. Furthermore, in the advanced tasks such as Phrase Detection and Relationship Detection, the AVR is significantly better than state-of-the-art methods by 10\%$\sim$21\% on recall. On the other hand, we also test AVR on the larger VG datasets (VG-VtransE \cite{Vtranse} and VG-MSDN \cite{MSDN}), and show the results  on Table \ref{tab:cmp_vg}. It can be observed that the Baseline with the priori module can offer better results than the original Baseline on three tasks for all $K$. Meanwhile, the Baseline combined with $\rm Priori_{rw}$ and attention module ($\rm Baseline$+$\rm Priori_{rw}$+$\rm Att$) has significantly improvement compared with the other methods. In some difficult cases, e.g., in the Relationship Detection task on the VG-VtransE dataset, the improvement can be up to $87.5\%$ (marked with underline in the table). 

\begin{table*}[tb]
    \small
    \centering
    \renewcommand\tabcolsep{5.0pt}
    \begin{tabular}{cccccccc}
    \toprule
    \multirow{2}{*}{K} & \multirow{2}{*}{Method} & \multicolumn{2}{c}{Predicate Det.} & \multicolumn{2}{c}{Phrase Det.} & \multicolumn{2}{c}{Relaitonship Det.} \\
    & & Rec@50 & Rec@100 & Rec@50 & Rec@100 & Rec@50 & Rec@100 \\
    \midrule
    \multirow{9}{*}{1}   & LP \cite{LangPrior}           & 47.87 & 47.87 & 16.17 & 17.03 & 13.86 & 14.70 \\
                         & VtransE \cite{Vtranse}         & 44.76 & 44.76 & 19.42 & 22.42 & 14.07 & 15.20 \\
                         & PPR-FCN \cite{PPR-FCN}        & 47.43 & 47.43 & 19.62 & 23.15 & 14.41 & 15.72 \\
                         & ViP-CNN \cite{ViP-CNN}        & -     & -     & 22.78 & 27.91 & 17.32 & 20.01 \\
                         & LK \cite{LKDistill}           & 55.16 & 55.16 & 23.14 & 24.03 & 19.17 & 21.34 \\
                         & CAI \cite{CAI}                & 53.59 & 53.59 & 17.60 & 19.24 & 15.63 & 17.39 \\
                         & CAI+SCA-M \cite{Zoomnet}      & \textbf{55.98} & \textbf{55.98} & 25.21 & 28.89 & 19.54 & 22.39 \\
                         & F-Net \cite{FactorizableNet}  & -     & -     & 26.03 & 30.77 & 18.32 & 21.20 \\
                         & our AVR                       & 55.61 & 55.61 & \textbf{29.33} & \textbf{33.27} & \textbf{22.83} & \textbf{25.41} \\
    \midrule
    \multirow{5}{*}{70}  & LK \cite{LKDistill}           & 85.64 & 94.65 & 26.32 & 29.43 & 22.68 & 31.89 \\
                         & DR-NET \cite{DRnet}           & 80.78 & 81.90 & 19.93 & 23.45 & 17.73 & 20.88 \\
                         & DSR \cite{StructuralRanking}  & 86.01 & 93.18 & -     & -     & 19.03 & 23.29 \\
                         & CAI+SCA-M \cite{Zoomnet}      & 89.03 & 94.56 & 29.64 & 38.39 & 22.34 & 28.52 \\
                         & our AVR                       & \textbf{90.73} & \textbf{95.72} & \textbf{34.51} & \textbf{41.36} & \textbf{27.35} & \textbf{32.96} \\
    \bottomrule
    \end{tabular}
    \caption{Comparison of AVR and state-of-the-art methods on VRD dataset.}
    \label{tab:cmp_vrd}
\end{table*}

\begin{table*}[tb]
    \small
    \centering
    \begin{tabular}{c|cccccccc}
    \toprule
    \multirow{2}{*}{Dataset} & \multirow{2}{*}{K} & \multirow{2}{*}{Method} & \multicolumn{2}{c}{Predicate} & \multicolumn{2}{c}{Phrase} & \multicolumn{2}{c}{Relaitonship} \\
       & & & Rec@50 & Rec@100 & Rec@50 & Rec@100 & Rec@50 & Rec@100 \\
    \midrule
    \multirow{8}{*}{VG-VtransE}   & \multirow{5}{*}{1}   & VtransE \cite{Vtranse}           & 62.63 & 62.87 & 9.46  & 10.45 & 5.52  & 6.04  \\
                      &   & Fo+L \cite{structuredLearning}   & -     & -     & 13.07 & 15.61 & 6.82  & \underline{8.00}  \\
                      &   & Baseline                             & 70.45 & 70.46 & 13.65 & 17.69 & 7.21  & 9.20  \\
                      &   & Baseline+$\rm Priori_{rw}$            & \textbf{71.65} & \textbf{71.65} & 19.75 & 24.20 & 10.72 & 13.19 \\
                      &   & Baseline+$\rm Priori_{rw}$+Att        & -     & -     & \textbf{23.08} & \textbf{27.01} & \textbf{12.79} & \textbf{\underline{15.06}} \\
    \cline{2-9}
       & \multirow{3}{*}{100}  & Baseline                             & 96.17 & 98.59 & 14.13 & 18.75 & 7.60  & 10.15 \\
                      &   & Baseline+$\rm Priori_{rw}$            & \textbf{96.89} & \textbf{99.01} & 20.07 & 24.98 & 10.97 & 13.80 \\
                      &   & Baseline+$\rm Priori_{rw}$+Att        & -     & -     & \textbf{24.20} & \textbf{29.91} & \textbf{13.60} & \textbf{17.09} \\
    \midrule
    \multirow{8}{*}{VG-MSDN} & \multirow{5}{*}{1}   & ISGG \cite{MessagePassing}       & -     & -     & 15.87 & 19.45 & 8.23  & 10.88  \\
                      &   & MSDN \cite{MSDN}                 & -     & -     & 19.95 & 24.93 & 10.72 & 14.22 \\
                      &   & F-Net \cite{FactorizableNet}     & -     & -     & 22.84 & 28.57 & 13.06 & 16.47 \\
                      &   & Baseline                             & 62.92 & 62.93 & 20.08 & 24.62 & 11.94 & 14.60 \\
                      &   & Baseline+$\rm Priori_{rw}$            & \textbf{64.97} & \textbf{64.97} & 23.70 & 29.12 & 14.50 & 17.82 \\
                      &   & Baseline+$\rm Priori_{rw}$+Att        & -     & -     & \textbf{29.12} & \textbf{32.29} & \textbf{18.33} & \textbf{19.97} \\
    \cline{2-9}
       & \multirow{3}{*}{50}  & Baseline                             & 93.81 & 97.87 & 20.67 & 26.01 & 12.44 & 15.81 \\
                      &   & Baseline+$\rm Priori_{rw}$            & \textbf{93.97} & 97.63 & 23.99 & 30.11 & 14.72 & 18.61 \\
                      &   & Baseline+$\rm Priori_{rw}$+Att        & -     & -     & \textbf{31.27} & \textbf{37.91} & \textbf{19.95} & \textbf{24.24} \\
    \bottomrule
    \end{tabular}
    \caption{Comparison of AVR and state-of-the-art methods on VG datasets.}
    \label{tab:cmp_vg}
\end{table*}

\begin{figure*}[t]
    \centering
    \includegraphics[width=\linewidth]{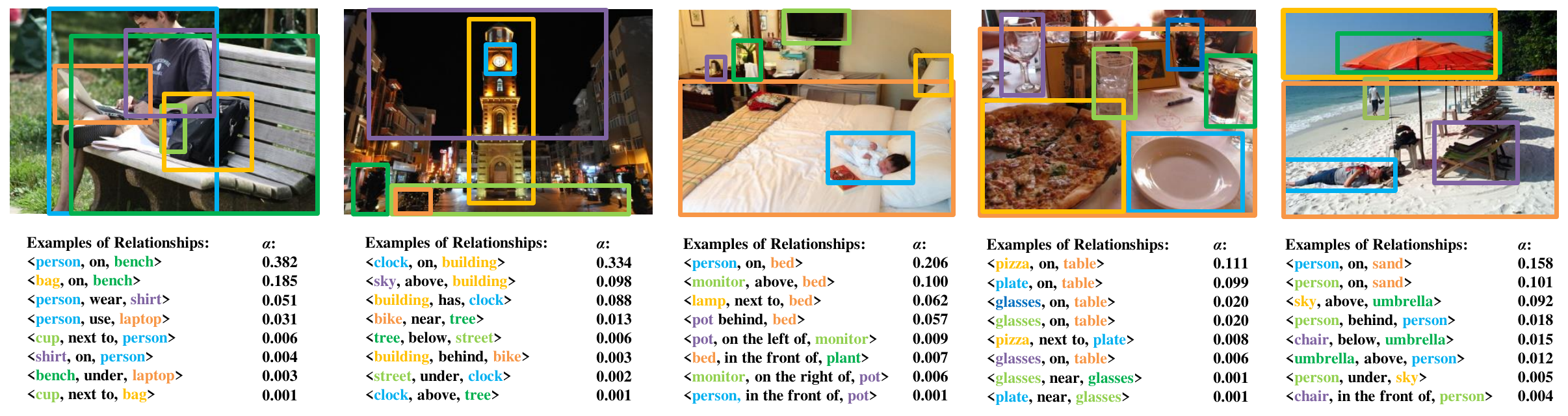}
    \caption{Qualitative examples of detected relationships and the salient weights $\alpha$ given by the proposed AVR. The objects in colors are localized with the bounding boxes with same colors.}
    \label{fig:example1}
\end{figure*}

Furthermore, to show the performance of the AVR model intuitively, we visualize some test examples with the salient weights of predicted relationships in Fig.~\ref{fig:example1}. We can see that the predicates between objects are predicted accurately and the salient weights $ \alpha $  can precisely indicate the importance of the relationships in an image. For example, in the first image of Fig. \ref{fig:example1}, the salient weights of the relationships $<$person, on, bench$>$ and $<$bag, on, bench$>$ are larger than the weights of  $<$cup, next to, person$>$ and $<$shirt, on, person$>$. This indicates AVR can pay more attention on the salient relationships in the image. 

\begin{figure}[t]
    \centering
    \includegraphics[width=\linewidth]{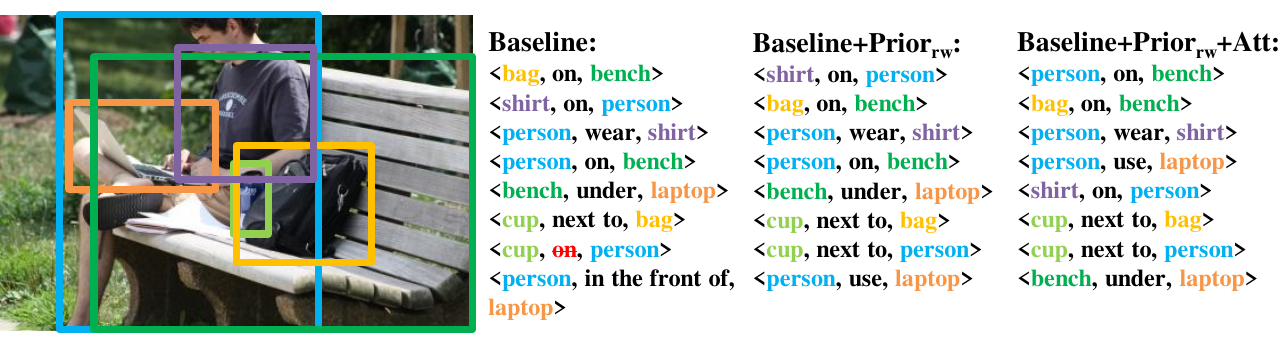}
    \caption{According the confident probabilities of relationships, the several sorting relationships of three different models (i.e., Baseline, Baseline+$\rm Priori_{rw}$ and Baseline+$\rm Priori_{rw}$+Att) are listed for comparison.}
    \label{fig:example2}
\end{figure}

To further visualize the effectiveness of different components of AVR, we show the ranking results of  three variation models (i.e., Baseline, Baseline+$\rm Priori_{rw}$ and Baseline+$\rm Priori_{rw}$+Att) on a test example in Fig. \ref{fig:example2}. It can be observed that the model with the attention module can significantly enhance the rankings of the important relationships (e.g., $<$person, on, bench$>$) and decrease the rankings of  relatively unimportant relationships (e.g., $<$cup, next to, bag$>$). Furthermore, the $\rm Priori_{rw}$ can effectively correct some mistakes about predicates, e.g., the $<$cup, on, person$>$ is revised to $<$cup, next to, person$>$, which is more precise to describe the relationship.

\section{Conclusion}\label{sec:conclusion}

In this paper, we propose an attention based model AVR to solve the problem of salient visual relationship detection. AVR can not only precisely recognize the interaction between objects by fusing visual, spatial and semantic features, but also distinguish the importance of relationships with the attention based mechanism. Besides the multi-view information in the local context of an image, the global context in the whole dataset is also used to improve the accuracy of predicate prediction by performing random walk on the heterogeneous priori knowledge graph. Comprehensive experiments are conducted on the real-world VRD and VG datasets and the results show that AVR outperforms state-of-the-art methods significantly.

%
% The acknowledgments section is defined using the "acks" environment (and NOT an unnumbered section). This ensures
% the proper identification of the section in the article metadata, and the consistent spelling of the heading.
% \begin{acks}
% \end{acks}

%
% The next two lines define the bibliography style to be used, and the bibliography file.
\bibliographystyle{aaai}
\bibliography{reference}

\begin{thebibliography}{}

\bibitem[\protect\citeauthoryear{Dai, Zhang, and Lin}{2017}]{DRnet}
Dai, B.; Zhang, Y.; and Lin, D.
\newblock 2017.
\newblock Detecting visual relationships with deep relational networks.
\newblock In {\em 2017 {IEEE} Conference on Computer Vision and Pattern
  Recognition, {CVPR} 2017, Honolulu, HI, USA, July 21-26, 2017},  3298--3308.

\bibitem[\protect\citeauthoryear{David and Fill}{1995}]{RandomWalk}
David, A., and Fill, J.
\newblock 1995.
\newblock Reversible markov chains and random walks on graphs.

\bibitem[\protect\citeauthoryear{Johnson \bgroup et al\mbox.\egroup
  }{2015}]{ImageRetrieval}
Johnson, J.; Krishna, R.; Stark, M.; Li, L.; Shamma, D.~A.; Bernstein, M.~S.;
  and Li, F.
\newblock 2015.
\newblock Image retrieval using scene graphs.
\newblock In {\em {IEEE} Conference on Computer Vision and Pattern Recognition,
  {CVPR} 2015, Boston, MA, USA, June 7-12, 2015},  3668--3678.

\bibitem[\protect\citeauthoryear{Krishna \bgroup et al\mbox.\egroup
  }{2017}]{VGDataset}
Krishna, R.; Zhu, Y.; Groth, O.; Johnson, J.; Hata, K.; Kravitz, J.; Chen, S.;
  Kalantidis, Y.; Li, L.; Shamma, D.~A.; Bernstein, M.~S.; and Fei{-}Fei, L.
\newblock 2017.
\newblock Visual genome: Connecting language and vision using crowdsourced
  dense image annotations.
\newblock {\em International Journal of Computer Vision} 123(1):32--73.

\bibitem[\protect\citeauthoryear{Li \bgroup et al\mbox.\egroup
  }{2017a}]{ViP-CNN}
Li, Y.; Ouyang, W.; Wang, X.; and Tang, X.
\newblock 2017a.
\newblock Vip-cnn: Visual phrase guided convolutional neural network.
\newblock In {\em 2017 {IEEE} Conference on Computer Vision and Pattern
  Recognition, {CVPR} 2017, Honolulu, HI, USA, July 21-26, 2017},  7244--7253.

\bibitem[\protect\citeauthoryear{Li \bgroup et al\mbox.\egroup }{2017b}]{MSDN}
Li, Y.; Ouyang, W.; Zhou, B.; Wang, K.; and Wang, X.
\newblock 2017b.
\newblock Scene graph generation from objects, phrases and region captions.
\newblock In {\em {IEEE} International Conference on Computer Vision, {ICCV}
  2017, Venice, Italy, October 22-29, 2017},  1270--1279.

\bibitem[\protect\citeauthoryear{Li \bgroup et al\mbox.\egroup
  }{2018}]{FactorizableNet}
Li, Y.; Ouyang, W.; Zhou, B.; Shi, J.; Zhang, C.; and Wang, X.
\newblock 2018.
\newblock Factorizable net: an efficient subgraph-based framework for scene
  graph generation.
\newblock In {\em Proceedings of the European Conference on Computer Vision
  (ECCV)},  335--351.

\bibitem[\protect\citeauthoryear{Liang \bgroup et al\mbox.\egroup
  }{2018}]{StructuralRanking}
Liang, K.; Guo, Y.; Chang, H.; and Chen, X.
\newblock 2018.
\newblock Visual relationship detection with deep structural ranking.
\newblock In {\em Thirty-Second AAAI Conference on Artificial Intelligence}.

\bibitem[\protect\citeauthoryear{Lu \bgroup et al\mbox.\egroup
  }{2016}]{LangPrior}
Lu, C.; Krishna, R.; Bernstein, M.; and Li, F.~F.
\newblock 2016.
\newblock Visual relationship detection with language priors.

\bibitem[\protect\citeauthoryear{Lu \bgroup et al\mbox.\egroup }{2018}]{R-VQA}
Lu, P.; Ji, L.; Zhang, W.; Duan, N.; Zhou, M.; and Wang, J.
\newblock 2018.
\newblock {R-VQA:} learning visual relation facts with semantic attention for
  visual question answering.
\newblock In {\em Proceedings of the 24th {ACM} {SIGKDD} International
  Conference on Knowledge Discovery {\&} Data Mining, {KDD} 2018, London, UK,
  August 19-23, 2018},  1880--1889.

\bibitem[\protect\citeauthoryear{Mikolov \bgroup et al\mbox.\egroup
  }{2013}]{word-vector}
Mikolov, T.; Chen, K.; Corrado, G.; and Dean, J.
\newblock 2013.
\newblock Efficient estimation of word representations in vector space.
\newblock In {\em 1st International Conference on Learning Representations,
  {ICLR} 2013, Scottsdale, Arizona, USA, May 2-4, 2013, Workshop Track
  Proceedings}.

\bibitem[\protect\citeauthoryear{Pennington, Socher, and Manning}{2014}]{glove}
Pennington, J.; Socher, R.; and Manning, C.~D.
\newblock 2014.
\newblock Glove: Global vectors for word representation.
\newblock In {\em Empirical Methods in Natural Language Processing (EMNLP)},
  1532--1543.

\bibitem[\protect\citeauthoryear{Redmon and Farhadi}{2017}]{YOLO9000}
Redmon, J., and Farhadi, A.
\newblock 2017.
\newblock {YOLO9000:} better, faster, stronger.
\newblock In {\em 2017 {IEEE} Conference on Computer Vision and Pattern
  Recognition, {CVPR} 2017, Honolulu, HI, USA, July 21-26, 2017},  6517--6525.

\bibitem[\protect\citeauthoryear{Ren \bgroup et al\mbox.\egroup
  }{2017}]{Faster-RCNN}
Ren, S.; He, K.; Girshick, R.~B.; and Sun, J.
\newblock 2017.
\newblock Faster {R-CNN:} towards real-time object detection with region
  proposal networks.
\newblock {\em {IEEE} Trans. Pattern Anal. Mach. Intell.} 39(6):1137--1149.

\bibitem[\protect\citeauthoryear{Shen \bgroup et al\mbox.\egroup
  }{2018}]{RandomWalk-reid}
Shen, Y.; Li, H.; Xiao, T.; Yi, S.; Chen, D.; and Wang, X.
\newblock 2018.
\newblock Deep group-shuffling random walk for person re-identification.
\newblock In {\em 2018 {IEEE} Conference on Computer Vision and Pattern
  Recognition, {CVPR} 2018, Salt Lake City, UT, USA, June 18-22, 2018},
  2265--2274.

\bibitem[\protect\citeauthoryear{Wang \bgroup et al\mbox.\egroup
  }{2017}]{KnowledgeGraphEmb}
Wang, Q.; Mao, Z.; Wang, B.; and Guo, L.
\newblock 2017.
\newblock Knowledge graph embedding: {A} survey of approaches and applications.
\newblock {\em {IEEE} Trans. Knowl. Data Eng.} 29(12):2724--2743.

\bibitem[\protect\citeauthoryear{Xu \bgroup et al\mbox.\egroup
  }{2017}]{MessagePassing}
Xu, D.; Zhu, Y.; Choy, C.~B.; and Fei-Fei, L.
\newblock 2017.
\newblock Scene graph generation by iterative message passing.
\newblock In {\em Proceedings of the IEEE Conference on Computer Vision and
  Pattern Recognition},  5410--5419.

\bibitem[\protect\citeauthoryear{Yang, Zhang, and
  Cai}{2018}]{Shuffle-then-assemble}
Yang, X.; Zhang, H.; and Cai, J.
\newblock 2018.
\newblock Shuffle-then-assemble: learning object-agnostic visual relationship
  features.
\newblock In {\em Proceedings of the European Conference on Computer Vision
  (ECCV)},  36--52.

\bibitem[\protect\citeauthoryear{Yao \bgroup et al\mbox.\egroup
  }{2018}]{ImageCaptioning}
Yao, T.; Pan, Y.; Li, Y.; and Mei, T.
\newblock 2018.
\newblock Exploring visual relationship for image captioning.
\newblock In {\em Computer Vision - {ECCV} 2018 - 15th European Conference,
  Munich, Germany, September 8-14, 2018, Proceedings, Part {XIV}},  711--727.

\bibitem[\protect\citeauthoryear{Yin \bgroup et al\mbox.\egroup
  }{2018}]{Zoomnet}
Yin, G.; Sheng, L.; Liu, B.; Yu, N.; Wang, X.; Shao, J.; and Change~Loy, C.
\newblock 2018.
\newblock Zoom-net: Mining deep feature interactions for visual relationship
  recognition.
\newblock In {\em Proceedings of the European Conference on Computer Vision
  (ECCV)},  322--338.

\bibitem[\protect\citeauthoryear{Yu \bgroup et al\mbox.\egroup
  }{2017}]{LKDistill}
Yu, R.; Li, A.; Morariu, V.~I.; and Davis, L.~S.
\newblock 2017.
\newblock Visual relationship detection with internal and external linguistic
  knowledge distillation.
\newblock In {\em IEEE International Conference on Computer Vision}.

\bibitem[\protect\citeauthoryear{Zhang \bgroup et al\mbox.\egroup
  }{2017a}]{Vtranse}
Zhang, H.; Kyaw, Z.; Chang, S.-F.; and Chua, T.-S.
\newblock 2017a.
\newblock Visual translation embedding network for visual relation detection.
\newblock In {\em Proceedings of the IEEE conference on computer vision and
  pattern recognition},  5532--5540.

\bibitem[\protect\citeauthoryear{Zhang \bgroup et al\mbox.\egroup
  }{2017b}]{PPR-FCN}
Zhang, H.; Kyaw, Z.; Yu, J.; and Chang, S.
\newblock 2017b.
\newblock {PPR-FCN:} weakly supervised visual relation detection via parallel
  pairwise {R-FCN}.
\newblock In {\em {IEEE} International Conference on Computer Vision, {ICCV}
  2017, Venice, Italy, October 22-29, 2017},  4243--4251.

\bibitem[\protect\citeauthoryear{Zhu and Jiang}{2018}]{structuredLearning}
Zhu, Y., and Jiang, S.
\newblock 2018.
\newblock Deep structured learning for visual relationship detection.
\newblock In {\em Thirty-Second AAAI Conference on Artificial Intelligence}.

\bibitem[\protect\citeauthoryear{Zhuang \bgroup et al\mbox.\egroup
  }{2017}]{CAI}
Zhuang, B.; Liu, L.; Shen, C.; and Reid, I.
\newblock 2017.
\newblock Towards context-aware interaction recognition for visual relationship
  detection.
\newblock In {\em Proceedings of the IEEE International Conference on Computer
  Vision},  589--598.

\end{thebibliography}
% % 
% % If your work has an appendix, this is the place to put it.
% \appendix



% Lorem ipsum dolor sit amet, consectetur adipiscing elit. Morbi malesuada, quam in pulvinar varius, metus nunc fermentum urna, id sollicitudin purus odio sit amet enim. Aliquam ullamcorper eu ipsum vel mollis. Curabitur quis dictum nisl. Phasellus vel semper risus, et lacinia dolor. Integer ultricies commodo sem nec semper. 


% Etiam commodo feugiat nisl pulvinar pellentesque. Etiam auctor sodales ligula, non varius nibh pulvinar semper. Suspendisse nec lectus non ipsum convallis congue hendrerit vitae sapien. Donec at laoreet eros. Vivamus non purus placerat, scelerisque diam eu, cursus ante. Etiam aliquam tortor auctor efficitur mattis. 


% Nam id fermentum dui. Suspendisse sagittis tortor a nulla mollis, in pulvinar ex pretium. Sed interdum orci quis metus euismod, et sagittis enim maximus. Vestibulum gravida massa ut felis suscipit congue. Quisque mattis elit a risus ultrices commodo venenatis eget dui. Etiam sagittis eleifend elementum. 

% Nam interdum magna at lectus dignissim, ac dignissim lorem rhoncus. Maecenas eu arcu ac neque placerat aliquam. Nunc pulvinar massa et mattis lacinia.

\end{document}